\documentclass[10pt,twocolumn,letterpaper]{article}

\usepackage[accsupp]{axessibility}
\usepackage[pagenumbers]{cvpr}  

\usepackage{color,colortbl}
\NewDocumentCommand{\haoyu}{ mO{} }{\textcolor{red}{\textsuperscript{\textit{Haoyu}}\textsf{\textbf{\small[#1]}}}}

\NewDocumentCommand{\hg}{ mO{} }{\textcolor{blue}{\textsuperscript{\textit{Hg}}\textsf{\textbf{\small[#1]}}}}

\NewDocumentCommand{\yaqian}{ mO{} }{\textcolor{cyan}{\textsuperscript{\textit{yaqian}}\textsf{\textbf{\small[#1]}}}}

\usepackage{xspace}
\newcommand{\twodeval}{single-frame 2D segmentation\xspace}

\newcommand{\threedeval}{multi-frame 3D segmentation\xspace}

\usepackage{multicol, multirow}
\usepackage{tabularray}
\usepackage[ruled,vlined]{algorithm2e}

\title{Segment anything model 2: an application to 2D and 3D medical images}

\author{
 Haoyu Dong\textsuperscript{1}\thanks{Co-first authors}\text{ }, Hanxue Gu\textsuperscript{1}$^*$, Yaqian Chen\textsuperscript{1}, Jichen Yang\textsuperscript{1}, Yuwen Chen\textsuperscript{1}, Maciej A. Mazurowski\textsuperscript{1,2,3,4}\\
  \textsuperscript{1}Department of Electrical and Computer Engineering, Duke University \\
  \textsuperscript{2}Department of Radiology, Duke University \\
  \textsuperscript{3}Department of Biostatistics \& Bioinformatics, Duke University \\
  \textsuperscript{4}Department of Computer Science, Duke University \\
  {\tt\small{\{haoyu.dong151, hanxue.gu, maciej.mazurowski\}}@duke.edu} \\
}

\begin{document}
\maketitle

\begin{abstract}
Segment Anything Model (SAM) has gained significant attention because of its ability to segment various objects in images given a prompt. The recently developed SAM 2 has extended this ability to video inputs. This opens an opportunity to apply SAM to 3D images, one of the fundamental tasks in the medical imaging field. In this paper,
we extensively evaluate SAM 2's ability to segment both 2D and 3D medical images by first collecting 21 medical imaging datasets, including surgical videos, common 3D modalities such as computed tomography (CT), magnetic resonance imaging (MRI), and positron emission tomography (PET) as well as 2D modalities such as X-ray and ultrasound.
Two evaluation settings of SAM 2 are considered: (1) \threedeval, where prompts are provided to one or multiple slice(s) selected from the volume, and (2) \twodeval, where prompts are provided to each slice. The former only applies to videos and 3D modalities, while the latter applies to all datasets.
Our results show that SAM 2 exhibits similar performance as SAM under \twodeval, and has variable performance under \threedeval depending on the choices of slices to annotate, the direction of the propagation, the predictions utilized during the propagation, etc. 
We believe our work enhances the understanding of SAM 2's behavior in the medical field and provides directions for future work in adapting SAM 2 to this domain. Our code is available at: https://github.com/mazurowski-lab/segment-anything2-medical-evaluation.
\end{abstract}


%

\section{Introduction}
Medical image segmentation is crucial for multiple clinical applications such as disease diagnosis and clinical analysis \cite{lew2024publicly, patil2013medical, ramesh2021review, wang2024sam}. Despite advancements in medical imaging technologies, segmentation remains challenging due to the labor-intensive nature of data annotation and the complexity of medical images \cite{bran2024qubiq, mazurowski2023segment, yin2018medical}. 

Segment Anything Model (SAM) addresses these challenges in multiple directions. On the one hand, SAM has demonstrated impressive zero-shot segmentation performance with prompt inputs, significantly reducing the need for extensive manual data annotation \cite{cheng2023sam, mazurowski2023segment}. On the other hand, several works fine-tune SAM to specific tasks and demonstrate improvements over standard segmentation techniques \cite{gu2024segmentanybone, fu2023effectiveness, li2024polyp, na2024segment, ma2024segment}, such as nn-UNet \cite{isensee2021nnu}.
Despite these advancements, SAM's limitation to 2D images restricts its applicability to scenarios that require three-dimensional understanding \cite{ravi2024sam2}. 

Following the release of SAM, some work attempted to address this challenge by introducing additional components to SAM to enable its 3D segmentation capability. For example, SAM3D \cite{bui2023sam3d} combines the SAM encoder with a lightweight 3D CNN decoder; 3DSAM-A \cite{gong20233dsam} modifies the original prompt encoder and mask decoder to operate in 3D; SAM-Med3D \cite{wang2023sammed3d} introduces an additional 3D convolution before the image encoder and replaces 2D positional encoding layers with 3D one. The recently introduced SAM 2 \cite{ravi2024sam2} solves this limitation fundamentally by extending the backbone of SAM to 3D. Specifically, SAM 2 proposes a memory bank that retains information from past predictions and allows it to make predictions on slices without prompts based on the information.
This feature motivates us to examine SAM 2's ability to segment 3D medical images since video segmentation can be transferred to 3D segmentation seamlessly, \textit{i.e.,} each slice can be treated as a frame. Note that we will use the terms ``slice'' and ``frame'' interchangeably throughout the paper.

\begin{figure*}[t]
    \centering
    \includegraphics[width=1\linewidth]{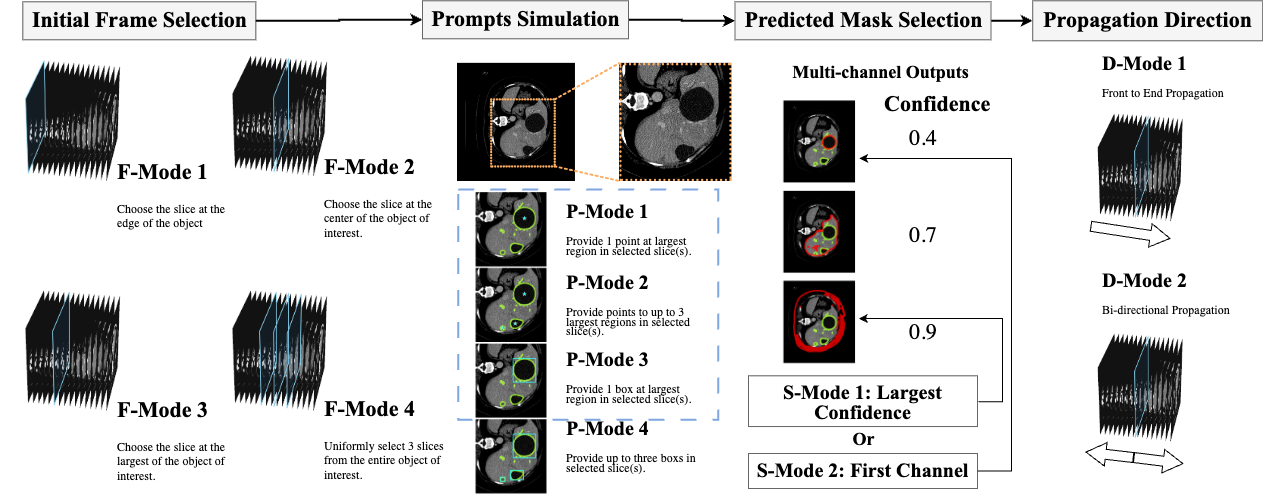}
    \caption{The pipeline of evaluating SAM 2 in the 3D setting. Different modes at each stage are proposed and evaluated.}
    \label{fig:pipeline}
\end{figure*}

In this paper, we extend the previous evaluation experiments on SAM \cite{mazurowski2023segment} to SAM 2, aiming to explore the model's effectiveness in a more complex, three-dimensional context. Specifically, we consider two evaluation settings: \threedeval and \twodeval. In \twodeval, prompts are provided to the object of interest on \textbf{each slice} \cite{mazurowski2023segment}, whereas in \threedeval, prompts are provided on \textbf{one or a few slices} selected from the volume. We further consider some unique challenges in \threedeval, such as the selection of the slice(s) to be annotated, the direction of propagation, \textit{i.e.,} predicting slices without prompts based on ones with prompts, and the selection of prediction during propagation. Our experiments are conducted across 21 datasets, covering 5 modalities (magnetic resonance imaging (MRI), computed tomography (CT), positron emission tomography (PET), X-ray, and Ultrasound) across different body locations and 3 different types of surgical videos. We observe several trends of SAM 2 under both evaluation pipelines and summarize them in Section \ref{sec:dis}.


\section{Methods}
\label{sec:method}

In this section, we discuss the two evaluation settings, \twodeval and \threedeval, in detail. 
Intersection over Union (IoU) is the evaluation metric throughout the paper. To have a comparable performance between 2D and 3D segmentation, IoU is only computed over non-empty slices.

\subsection{Evaluation Criteria for Single-Frame 2D Segmentation}\label{method single-frame}
During \twodeval, SAM 2 exhibits the same behavior as SAM in segmenting the object of interest based on prompts.
For datasets with 2D modalities, we run SAM 2 naturally at the image level. For datasets with 3D modalities, we simulate prompts for every slice of the volume.
Following previous work \cite{mazurowski2023segment}, we design \twodeval in a non-iterative manner in which all prompts are determined without feedback from any prior predictions. 
Specifically, the following four 2D prompting modes (P-Mode) are used:
\begin{enumerate}
    \item P-Mode 1: One point prompt placed at the center of the largest connected region of the object of interest.
    \item P-Mode 2: One point prompt placed at each separate connected region of the object (up to three points).
    \item P-Mode 3: One box prompt placed at the center of the largest connected region of the object of interest.
    \item P-Mode 4: One box prompt placed at each separate connected region of the object (up to three boxes).
\end{enumerate}
These modes present common prompting strategies used during interactive segmentation.



\begin{figure*}[t]
    \centering
    \includegraphics[width=\linewidth]{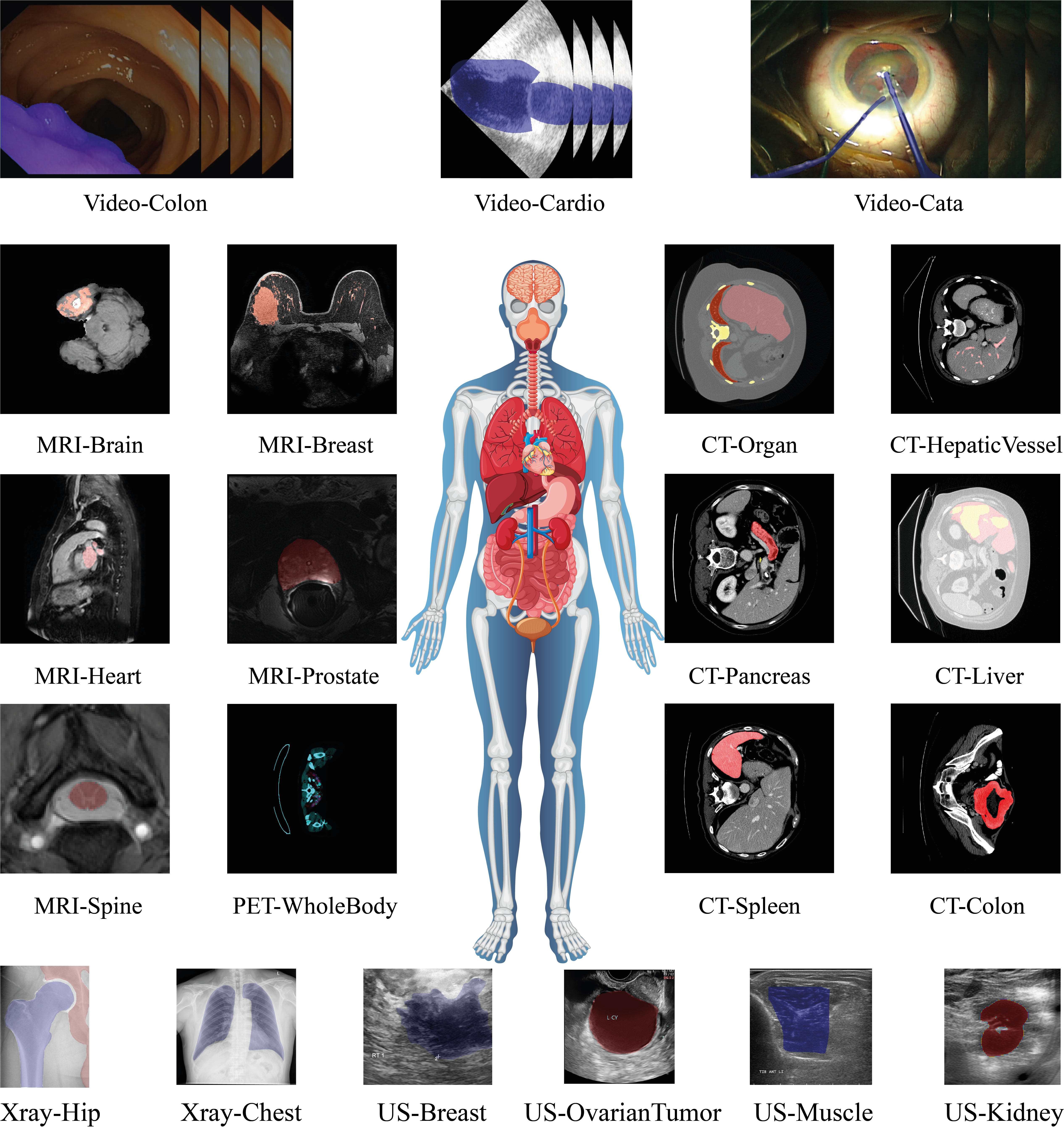}
    \caption{Examples from all 21 datasets, each overlaid with annotation masks. The top rows feature 15 examples from 3D datasets, while the bottom row presents 6 examples from 2D datasets. The human anatomy figure is from $Vecteezy.com$}
    \label{fig:dataset}
\end{figure*}

\subsection{Evaluation Criteria for Multi-Frame 3D Segmentation}
\label{sec:eval_3d}
SAM 2 differs from SAM mostly in its new ability to segment videos, which can be seamlessly transferred to the 3D image segmentation task.
In this section, we are mostly interested in SAM 2's semi-supervised segmentation ability, where we only provide prompts for one or a few frames in advance and use SAM 2 to predict other frames. The pipeline of evaluating SAM 2 in the \threedeval is shown in Figure \ref{fig:pipeline}, and we detail the choice for each stage next.

\noindent\textbf{Initial Frame Selection}. In video segmentation, the initial frame to be annotated is typically the first frame of video as it arrives first in the time stream and the object of interest usually does not change shape or size dramatically between consecutive frames \cite{wang2021unidentified,pont20172017}.
This is also the strategy used by SAM 2 for the semi-supervised video object segmentation task. 
However, such a strategy may be ineffective in 3D medical imaging because, in most scenarios, the boundary of the object of interest appears first, and the change between different frames can be significant. 
Instead, when annotating medical volumes, annotators tend to start with the slice with the object of interest being the most salient.
To evaluate SAM 2's original frame selection (choosing the first slice) and to approximate human behavior in annotation medical volumes, we consider the following four frame modes (F-Modes):
\begin{enumerate}
    \item F-Mode 1: Choose the edge slice when the object of interest first appears.
    \item F-Mode 2: Choose the slice at the center of the object of interest.
    \item F-Mode 3: Choose the slice with the largest object of interest.
    \item F-Mode 4: Uniformly select 3 slices from the entire object of interest.
\end{enumerate}

\noindent\textbf{Prompts Simulation}. When the slice to be annotated is selected, we utilize all but P-Mode 4 since multi-box prompts are not supported by SAM 2. In addition, we consider feeding ground truth (GT) masks as an additional prompt mode, denoted as P-Mode 5. This mode simulates the scenario of annotating one slice in the volume and letting SAM 2 predict the rest. 

\noindent\textbf{Propagation Direction}. After the first frame is annotated, SAM 2 can predict other non-conditioned frames (frames without prompts) through propagation. 
The propagation step utilizes the newly introduced memory structures that consist of a memory attention module and a memory encoder. The memory attention module merges the current slice's features with features from previously predicted slices and prompts and allows the network to make predictions on the current slice. The memory encoder compresses current predictions into a memory bank for future use. 




Note that since new prediction relies on the output from existing predictions, the order of propagation matters. 
Although the order of propagation might have less impact on video predictions in which the consecutive frames are similar and the object of interest appears consistently from beginning to end, it can have a more significant impact on 3D medical volumes in which the change between slices can be large and the object of interest does not occupy the full volume.
To take the order of propagation into account, we consider the following two direction modes (D-Mode):
\begin{enumerate}
    \item D-Mode 1 (Front-to-End Propagation): Start from the first slice of the volume and propagate forward through the volume.
    \item D-Mode 2 (Bi-directional Propagation): Start from the first annotated slice and propagate backward to the first slice; then restart from the annotated slice and propagate forward to the end of the volume.
\end{enumerate}
The ``restart'' in D-Mode 2 means predictions from the backward phase are not used during the forward propagation. We empirically find this to be beneficial for the final performance.

\noindent\textbf{Predicted Mask Selection}. Due to the ambiguity of providing point prompts, it is common to predict various levels of objects based on a single or a few point prompts. For example, if we place a point prompt on a tumor inside the brain, the object of interest can either be the tumor or the whole brain.
In the 2D setting of both SAM and SAM 2, the mask predictor can create multiple mask outputs from which humans can select the one closest to their needs. If there is no human involvement in the selection step, we can simply pick the first prediction channel, which represents the smallest area among the three channels based on previous empirical experiments \cite{mazurowski2023segment}. Alternatively, we can select the channel with the highest confidence based on the IoU calculated by SAM's mask decoder.

In 3D mode, to prevent propagating ambiguity across frames, SAM 2 chooses to pick the prediction with the highest confidence and only utilizes this prediction when estimating the object of interest on other frames. However, based on previous experiments with SAM \cite{mazurowski2023segment, gu2024build}, selecting the predictions with the highest confidence can be sub-par when using point prompts. This observation motivates us with the following predicted mask selection strategies:
\begin{enumerate}
    \item S-Mode 1: Selecting the predicted mask with the largest estimated IoU, \textit{i.e.,} confidence. 
    \item S-Mode 2: Selecting the first channel.
\end{enumerate}
Note that the selection of the first channel is not a typical hyperparameter tuning choice, as it requires modifications of SAM 2's inner structure. Therefore, we considered S-Mode 2 as a potential improvement of SAM 2 rather than an out-of-the-box configuration. \\

In summary, under \threedeval, we cover 4 slice selection modes, 4 prompt-simulation modes (P-Mode 1-3 plus P-Mode 5), 2 predicted mask strategies, and 2 propagation modes, resulting in 64 different experimental configurations in total. We evaluate the performance of all settings.

\begin{algorithm}[t]
\label{alg:1}
\caption{Correction-based Interactive prompting for Multi-Frame Segmentation}
\KwIn{Slices \( S = \{s_1, s_2, \dots, s_n\} \), \\
Sub-volume 1: \( S_{\text{backward}} \), Sub-volume 2: \( S_{\text{forward}} \), \\
Initial Prompts Queue: \( P_{\text{init}} = \{\} \), \\
SAM 2' Video Predictor: Predictor, \\
Number of Interactive Loops: \( K \)}
\KwOut{Final Predictions \( \text{Pred}_{\text{final}} \)}

\textbf{Step 1.} Initialize \( \text{Pred}_{\text{final}} = \{\} \)\;

\( P_{\text{init}} \gets \text{Frame\_Prompt\_Gen}(S, \text{Fmode}, \text{Pmode}) \)\;

\( P_{loop} \gets P_{\text{init}} \)\;

\textbf{Step 2.} \For{loop = 1 to K}{
    
    \( \text{pred}_{s_{\text{init}}} = \text{Predictor}(P_{\text{loop}}) \)\;
    
   \For{\( i = s_{\text{init}} \) \text{ to } 1}{
        \( \text{Pred}_{s_{\text{i}}} \gets \text{Predictor}(s_{\text{i}}) \)\;
    
        Save \( \text{Pred}_{s_{\text{min}}} \) in \( \text{Pred}_{\text{final}} \)\;
    }

    Identify the slice \( s_{\text{min}} \) in \( S_{\text{backward}} \) with the lowest IoU (within 16 slices of annotated slices)\;
        
    \( p_{\text{new}} \gets \text{GeneratePrompt}(s_{\text{min}}) \)\;
    
     \( P_{loop} \gets P_{\text{new}} \)\;
}

\textbf{Step 3.} Reinitialize Predictor\;

\( P_{loop} \gets P_{\text{init}} \)\;

\textbf{Step 4.} \For{loop = 1 to K}{
    
   \For{\( i = s_{\text{init}} \) \text{ to } end}{
        \( \text{Pred}_{s_{\text{min}}} \gets \text{Predictor}(s_{\text{min}}) \)\;
        
        Save \( \text{Pred}_{s_{\text{min}}} \) in \( \text{Pred}_{\text{final}} \)\;
    }
        Identify the slice \( s_{\text{min}} \) in \( S_{\text{forward}} \) with the lowest IoU (within 16 slices of annotated slices)\;
        
        \( p_{\text{new}} \gets \text{GeneratePrompt}(s_{\text{min}}) \)\;
        
        \( P_{loop} \gets P_{\text{new}} \)\;
        
}

\Return \( \text{Pred}_{\text{final}} \)\;

\end{algorithm}

\begin{algorithm}[t]
\label{alg:2}
\caption{Reinitialization-based Interactive Prompting for Multi-Frame Segmentation}
\KwIn{Slices \( S = \{s_1, s_2, \dots, s_n\} \), \\ 
Initial Prompts Queue: \( P_{\text{init}} = \{\} \), \\ 
SAM 2' Video Predictor: Predictor, \\ 
Number of Interactive Loops: \( K \), \\
Frame mode and Prompt mode: Fmode, Pmode}
\KwOut{Final Predictions \( \text{Pred}_{\text{final}} \)}

Initialize \( \text{Pred}_{\text{final}} = \{\} \) \;

Initialize \( \text{loop} = 0 \) \;

\( P_{\text{init}} \gets \text{Frame\_Prompt\_Gen}(S, \text{Fmode}, \text{Pmode}) \)

\For{loop = 1 to K}{
    \textbf{Step 1.} Reinitialize Predictor\;
    
    \textbf{Step 2.} \( Pred_{s_\text{init}} = \text{Predictor}(P_{\text{init}}) \)\;

    \textbf{Step 3} \For{\( i = s_{\text{init}} \) \text{ to } 1}{
        \( \text{pred}_{s_i} \gets \text{Predictor}(s_i, P_{\text{init}}) \)\;
        
        Save \( \text{pred}_{s_i} \) in \( \text{Pred}_{\text{final}} \)\;
    }

     \textbf{Step 4.} Reinitialize Predictor\;
     
    \textbf{Step 5.} \( Pred_{s_\text{init}} = \text{Predictor}(P_{\text{init}}) \)\;
    
    \textbf{Step 6.} \For{\( i = s_{\text{init}} \) \text{ to } \text{end}}{
        \( \text{pred}_{s_i} \gets \text{Predictor}(s_i, P_{\text{init}}) \)\;
        
        Save \( \text{pred}_{s_i} \) in \( \text{Pred}_{\text{final}} \)\;
    }
    
    \textbf{Step 7.} Identify the slice \( s_{\text{max}} \) with the largest error in \( \text{Pred}_{\text{final}} \)\;
    
    \textbf{Step 8.} \( \text{FN}_{\text{max}} \gets \text{FindMaxFN}(s_{\text{max}}) \) {\# find largest false negative region}\;

    \textbf{Step 9.} \( p_{\text{new}} \gets \text{GeneratePrompt}(\text{FN}_{\text{max}}) \)\;
    
    \textbf{Step 10.} \( \text{Add } p_{\text{new}} \text{ to } P_{\text{init}} \)\;
}
\Return \( \text{Pred}_{\text{final}} \)\;

\end{algorithm}

\subsection{Evaluation Criteria for Interactive Multi-Frame 3D Segmentation}

\begin{table*}[t]
\centering
\begin{tabular}{|m{3cm}<{\centering}|m{6.3cm}<{\centering}|m{1.5cm}<{\centering}|m{1.0cm}<{\centering}|m{1.5cm}<{\centering}|m{1cm}<{\centering}|}
\toprule
\textbf{Abbreviated dataset name} & \textbf{Full dataset name and citation} & \textbf{Modality} & \textbf{Num. classes} & \textbf{Object(s) of interest} & \textbf{Num. masks} \\ 
\midrule
\midrule
Xray-Chest & Montgomery County and Shenzhen Chest X-ray Datasets \cite{jaeger2014two} & X-ray & 1 & Chest & 704 \\ 
\midrule
Xray-Hip & X-ray Images of the Hip Joints \cite{gut2021x} & X-ray & 2 & Ilium, Femur & 140 \\
\midrule
US-Breast & Dataset of Breast Ultrasound Images \cite{al2020dataset} & Ultrasound & 1 & Breast & 630 \\ 
\midrule
US-Kidney & CT2US for Kidney Segmentation \cite{song2022ct2us} & Ultrasound & 1 & Kidney & 4,586 \\
\midrule
US-Muscle & Transverse Musculoskeletal Ultrasound Image Segmentations \cite{marzola2021deep}& Ultrasound & 1 & Muscle & 4,044 \\ 
\midrule
US-Ovarian-Tumor & Multi-Modality Ovarian Tumor Ultrasound (MMOTU) \cite{zhao2022multi} & Ultrasound & 1 & Ovarian tumor & 1,469 \\ 
\bottomrule
\end{tabular}
\vspace{2mm}
\caption{\textbf{2D datasets evaluated in this paper:} “num. masks” refers to the number of images with non-zero masks.}
\label{table_2D_dataset}
\end{table*}

\begin{table*}[t]
\centering
\begin{tabular}{|m{2cm}<{\centering}|m{5.1cm}<{\centering}|m{1.6cm}<{\centering}|m{1cm}<{\centering}|m{3cm}<{\centering}|m{1cm}<{\centering}|m{1cm}<{\centering}|}
\toprule
\textbf{Abbreviated dataset name} & \textbf{Full dataset name and citation} & \textbf{Modality} & \textbf{Num. classes} & \textbf{Object(s) of interest} & \textbf{Num. masks} & \textbf{Num. volumes}\\ 
\midrule
\midrule
MRI-Spine & Spinal Cord Grey Matter Segmentation Challenge \cite{prados2017spinal} & MRI & 2 & Gray matter and spinal cord & 551 &40\\ 
\midrule
MRI-Heart & Medical Segmentation Decathlon \cite{simpson2019large} & MRI & 1 & Heart & 1,301 &20\\
\midrule
MRI-Prostate & Initiative for Collaborative Computer Vision Benchmarking \cite{lemaitre2015computer} & MRI & 1 & Prostate & 1854 &115\\ 
\midrule
MRI-Brain & The Multimodal Brain Tumor Image Segmentation Benchmark (BraTS) \cite{menze2014multimodal} & MRI & 3 & GD-enhancing tumor, Peritumoral edema, necrotic and non-enhancing tumor core & 12,591 &206\\ 
\midrule
MRI-Breast & Duke Breast Cancer MRI: Breast + FGT Segmentation\cite{saha2018machine, hu2022fully} & MRI & 2 & Vessel and fibroglandular tissue & 14438 &100\\
\midrule
CT-Colon & Medical Segmentation Decathlon \cite{simpson2019large} & CT & 1 & Colon cancer primaries & 1,285 &126\\ 
\midrule
CT-HepaticVessel & Medical Segmentation Decathlon \cite{simpson2019large} & CT & 1 & Vessels & 13,046 &303\\ 
\midrule
CT-Pancreas & Medical Segmentation Decathlon \cite{simpson2019large} & CT & 1 & Parenchyma and mass & 8,792 &281\\ 
\midrule
CT-Spleen & Medical Segmentation Decathlon \cite{simpson2019large}) & CT & 1 & Spleen & 1,051 &41\\ 
\midrule
CT-Liver & The Liver Tumor Segmentation Benchmark (LiTS) \cite{bilic2023liver} & CT & 1 & Liver & 5,501 &131\\ 
\midrule
CT-Organ & CT Volumes with Multiple Organ Segmentations (CT-ORG) \cite{rister2019ct} & CT & 5 & Liver, bladder, lungs, kidney, and bone & 4,776 &10\\ 
\midrule
PET-Whole-Body & A FDG-PET/CT dataset with annotated tumor lesions \cite{gatidis2022whole} & PET/CT & 1 & Lesion & 1,015 &42\\ 
\midrule
Video-Cata & Cataract Surgery Dataset \cite{ghamsarian2023cataract} & Video & 1 & Medical Device & 1,778 & 30\\ 
\midrule
Video-Colon & Endoscopic Vision Challenge \cite{bernal2015wm} & Endoscopic Video& 1 & Polyps & 612 & 29\\ 
\midrule
Video-Cardio & CAMUS-Human Heart Data \cite{leclerc2019deep} & Ultrasound Video & 1 & Heart & 19,232 & 1,000\\ 

\bottomrule
\end{tabular}
\vspace{2mm}
\caption{\textbf{3D datasets evaluated in this paper:} “num. masks” refers to the number of images with non-zero masks. For 2D segmentation models, slices are used as inputs.}
\label{table_3D_dataset}
\end{table*}

In the previous section, prompts are provided before the propagation step.
Now we consider the interactive setting where users can refine SAM 2's predictions. 
SAM 2 provides two scenarios in this setting: (1) \textit{offline} evaluation, where users can review the entire video multiple times to identify and correct the frames with the largest model errors; and (2) \textit{online} evaluation, where users can only view the frames once as the video plays, placing prompts on the first frame with an IoU below a specified threshold. 
Since we are adapting video segmentation techniques to the 3D medical imaging segmentation task, our focus will be on the \textit{offline} setting, which closely aligns with the human annotation pipeline, where users would review the volumes multiple times before annotating the next slice.

Following SAM 2's \textit{offline} setting, we aim to find the frames with the largest prediction error. However, in the 3D segmentation task, we find bidirectional propagation to be more effective than front-to-end propagation (as will be demonstrated in Section \ref{sec:exp_propagation}). This means we cannot fully follow the original interactive pipeline. In response to this, we propose a correction-based interactive prompting strategy (as shown in Algorithm 1) that approximates the original setting in the bidirectional mode. Specifically, we divide the input volume into two sub-volumes, separated by the annotated slice, and run the original interactive pipeline on each sub-volume.
Note that in this scenario, the next slice to be annotated has to be within 16 slices given the default configuration of SAM 2. 
We further propose a reinitialization-based interactive prompting strategy (as shown in Algorithm 2). It interactively find lowest IoU slice within the volume and adds new prompts to the initial prompts list and reinitializes the model at each iteration.

The key distinction between the two algorithms lies in their initialization: in Algorithm 1, once the slice with the lowest prediction accuracy is identified, we immediately add new prompts to correct this slice's prediction while preserving the memory of the entire predictor throughout each round of correction. In contrast, in Algorithm 2, we do not correct the prediction but let SAM 2 make predictions solely on the new prompts.

\section{Dataset}

Consistent with the previous experimental study on SAM \cite{mazurowski2023segment}, this study utilizes 21 diverse medical datasets to evaluate the performance of SAM 2. Specifically, all datasets are evaluated during \twodeval, and 15 datasets with 3D modalities (MRI, CT, PET, and videos) are evaluated during \threedeval. 
The pre-processing steps are consistent with the previous study, except that for SAM 2 3D mode, we have to convert the input images to JPG format instead of PNG. We also provided visual representations of the annotations masks for each dataset in Figure \ref{fig:dataset}.

\subsection{2D Datasets}\label{2D Datasets}
We included 6 2D datasets, 2 X-rays, and 4 ultrasounds, covering 7 different anatomical objects. Specifically, the X-ray datasets cover chest and hip joint segmentation, and the ultrasound datasets encompass a broader range of regions, including breast, kidney, muscle, and ovarian tumor segmentation. Detailed information on these datasets can be found in Table \ref{table_2D_dataset}. To keep the format consistent with 3D datasets, the images of 2D datasets are also converted to JPG using the same pipeline.

\subsection{3D Datasets}\label{3D datasets}
We included 15 3D datasets, 5 MRI, 6 CT, 1 PET-CT, and 3 videos, covering 20 different anatomical objects. The structures of the datasets are modified to fit the input requirements of SAM 2 3D, where each individual volume has its own folder, and each slice is numbered according to its position within the volume. Table \ref{table_3D_dataset} demonstrates more detailed information.

\begin{figure*}[t]
    \centering
    \includegraphics[width=\linewidth]{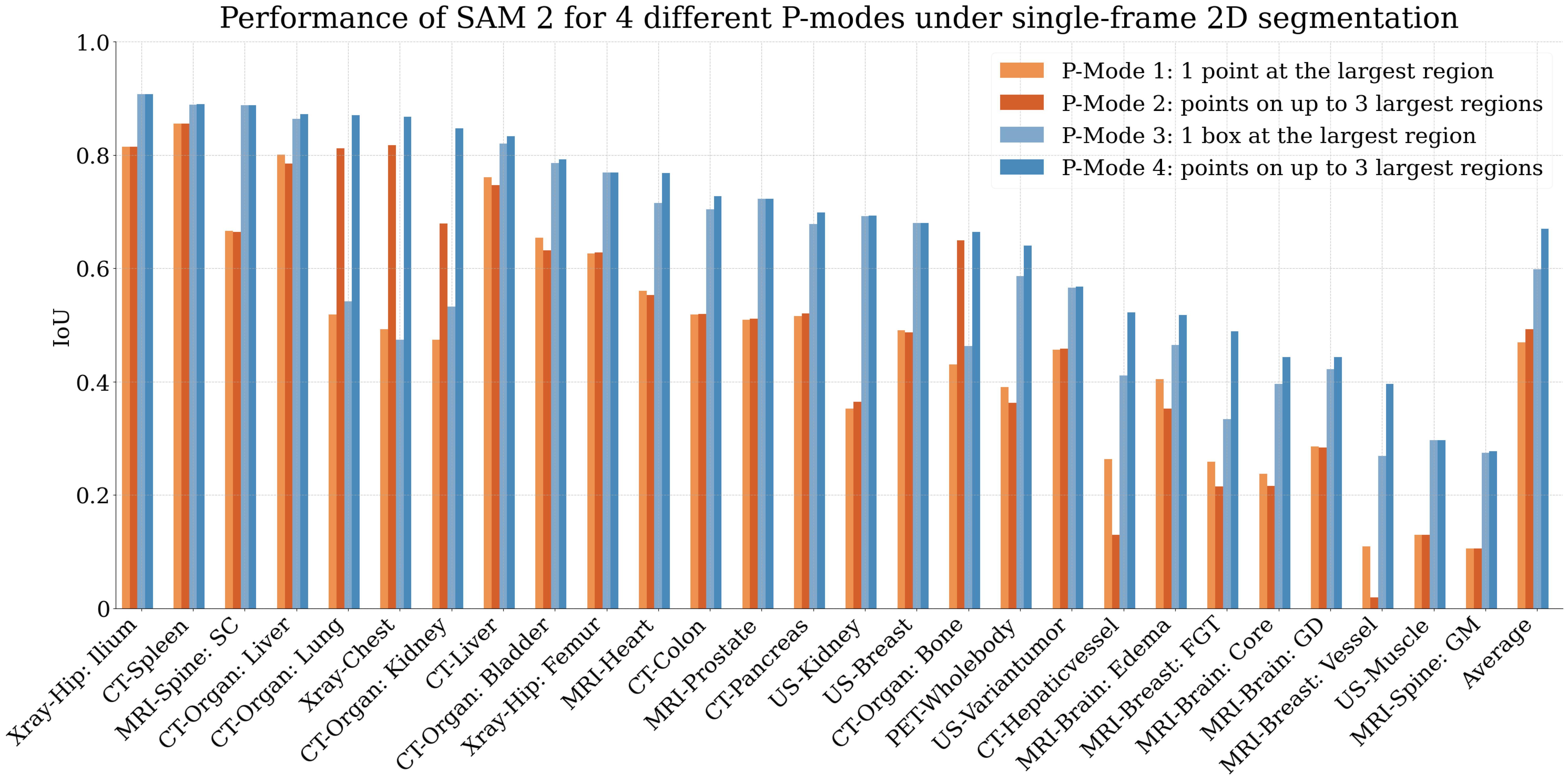}
    \caption{The performance of SAM 2 under \twodeval. Four prompt modes are considered, with results ranked in descending order based on P-Mode 4.}
    \label{fig:sam2_2d}
\end{figure*}

\begin{figure*}[t]
    \centering
    \includegraphics[width=\linewidth]{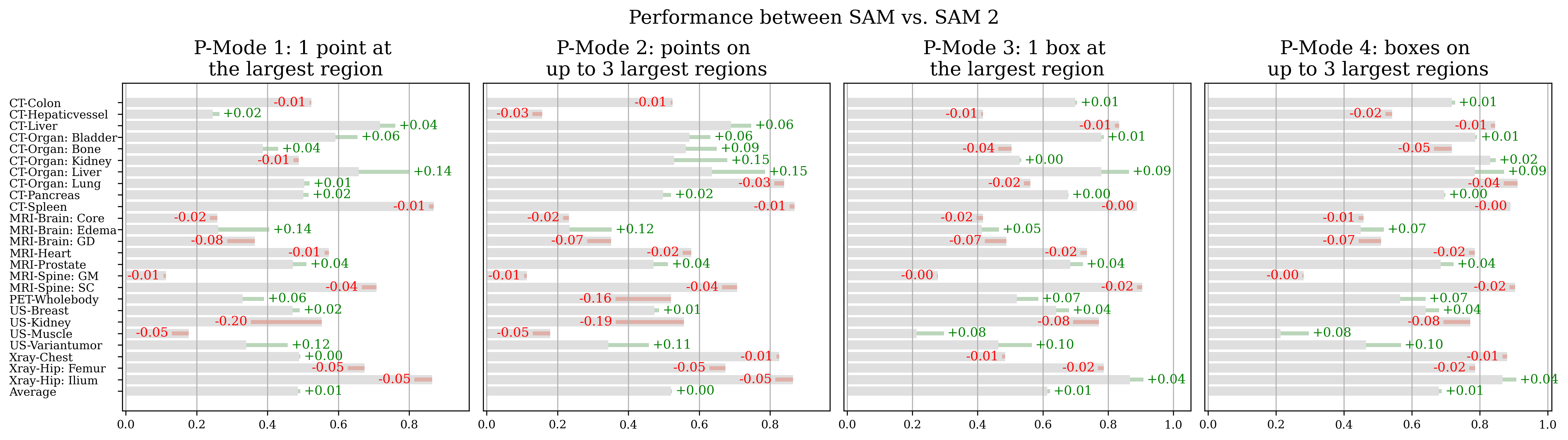}
    \caption{The single-frame 2D segmentation performance of SAM under 4 prompting modes 24 segmentation tasks (in gray) and the difference between the performance of SAM 2 and SAM. The differences are highlighted in red (when SAM has a higher IoU) and green (when SAM 2 has a higher IoU)}
    \label{fig:result_diff_2d}
\end{figure*}

\section{Experimental Results}
\subsection{Results of SAM 2 under Singe-frame 2D Segmentation}
In this section, we present the performance of SAM 2 under \twodeval scenario across four prompting modes (refer to Section \ref{method single-frame}). Figure \ref{fig:sam2_2d} shows the results for 2D and 3D datasets, respectively. The findings indicate that SAM 2's single-frame 2D segmentation capability is comparable with that of SAM (a comparison of single-frame 2D segmentation results between SAM and SAM 2 on individual datasets is provided in Figure \ref{fig:result_diff_2d}). Similar to SAM, the performance of SAM 2 varies significantly across different datasets. For instance, SAM 2 achieves an impressive IoU of 0.908 on the Xray-Hip dataset for ilium but performs poorly with an IoU of 0.278 on the MRI-Spine dataset for gray matter.

Comparing the performance for different prompting modes, we find box prompts consistently provide better results than point prompts. Moreover, providing more points does not always improve the performance. One explanation is that additional prompts are only provided when there is more than one disconnected region. In most medical image datasets, the target object usually appears as a single connected region. For datasets with multiple objects, such as the CT-Organ: Lung, supplying more positive prompts can assist the model in accurately segmenting both sides of the lung, thereby enhancing performance. Conversely, for smaller targets like those in the MRI-Brain dataset, providing additional prompts may cause the model to over-segment, potentially including the entire brain, which can lead to more significant error accumulation during propagation and be less effective than segmenting only a part of it with fewer prompts.

\begin{figure*}[t]
    \centering
    \includegraphics[width=\linewidth]{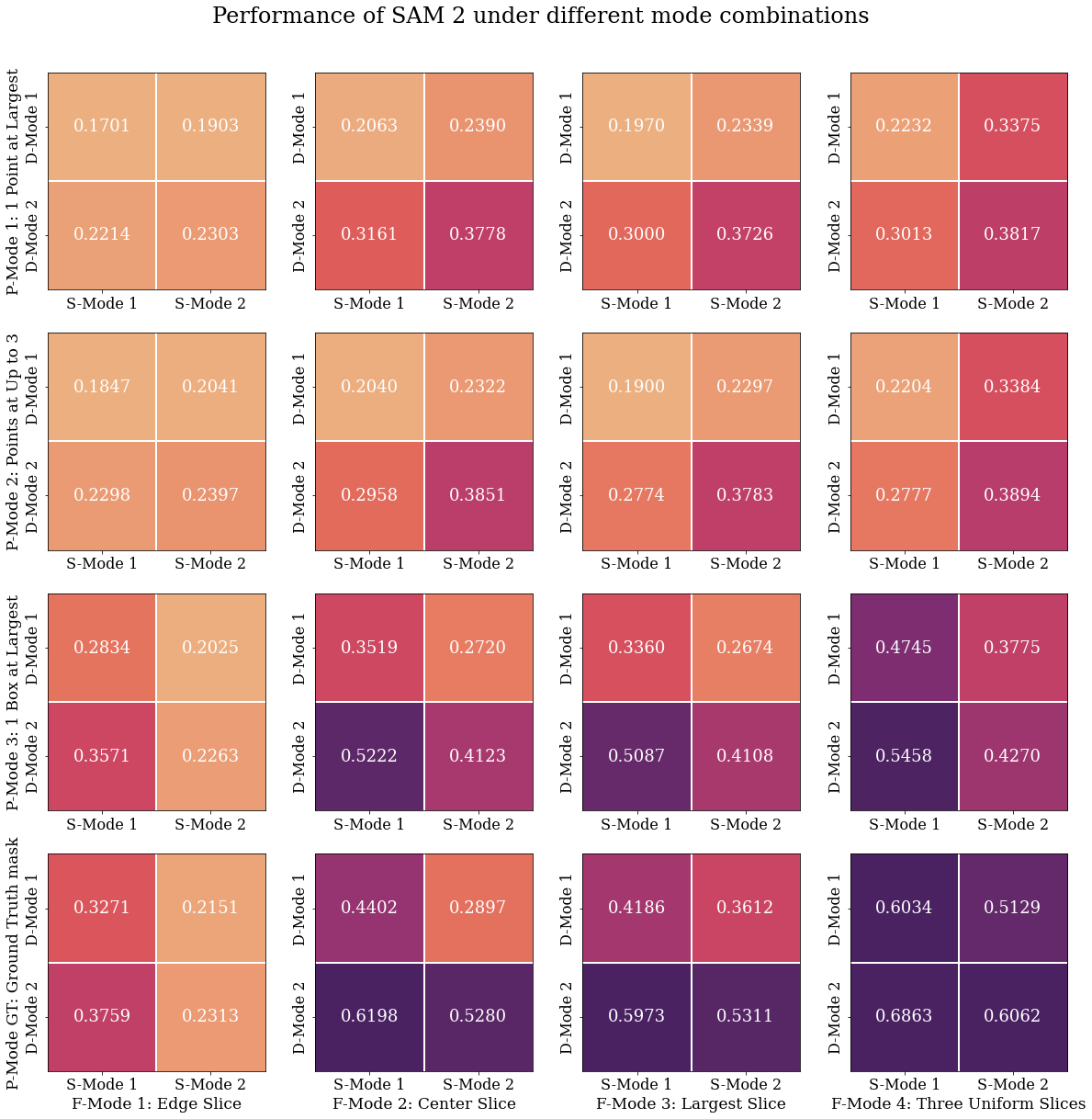}
    \caption{The multi-frame 3D segmentation performance of SAM 2 under all mode combinations, averaged over all datasets. \{F, P, S, D\}-Mode stands for the \textbf{f}rame to annotate, the \textbf{p}rompt type, \textbf{s}election of the predicted masks, and \textbf{d}irection of propagation respectively. The details of each model are shown in Figure \ref{fig:pipeline} and Sec. \ref{sec:method}.}
    \label{fig:result_enumerate}
\end{figure*}

\subsection{Results of SAM 2 under Multi-frame 3D Segmentation}
In this section, we investigate the impact of each component during \threedeval. The average performance of all mode combinations is shown in Figure \ref{fig:result_enumerate} and the performance on each dataset is shown in the Appendix.

\subsubsection{Impact of Propagation Mode}
\label{sec:exp_propagation}
The comparison between different rows in each small block in Figure \ref{fig:result_enumerate} shows the consistent superiority of bidirectional propagation. Namely, the improvements in average IoU are significant when starting from non-edge slices, with a minimal improvement of 0.0874 (F-Mode2, P-Mode2, S-Mode1) and a maximum improvement of 0.2383 (F-Mode4, P-Mode 5, S-Mode2). When starting from the edge slices, bi-directional propagation differs from front-to-end propagation in that the former begins at the annotated slice, while the latter starts at the first slice.

These findings suggest that, although SAM 2 supports prompting on any frame as a condition to predict new frames, the propagation is more effective when first predicting adjacent slices. The effectiveness of bidirectional propagation also shows that reverse propagation, \textit{i.e.,} predicting the (N-1)th slice based on the Nth slice, works well.

\subsubsection{Impact of Predicted Mask Selection}
To investigate the impact of prediction selection strategies, we can compare different columns in each small block in Figure \ref{fig:result_enumerate}. The results demonstrate that when using point prompts (P-Mode1 and P-Mode2), opting for the first channel's prediction achieves better performance than SAM 2's default choice of selecting the most confident prediction. In contrast, the opposite trend is observed when using box prompts (P-Mode 3) or ground truth masks (P-Mode 5).
One possible reason for the difference in trend is that the ambiguity is more severe when putting point prompts on objects. Box prompts are more definite and there is no ambiguity when providing ground truth masks. In these scenarios, selecting the most confident prediction is beneficial to the final performance.

\subsubsection{Impact of Initial Frame Selection}
To compare different choices of the initial frame selection, we can compare between different columns in Figure \ref{fig:result_enumerate}.
First, we observe that when only annotating a single slice (F-Mode 1-3), selecting the middle slice or the slice with the largest object of interest yields similar performance, while selecting the edge slice gives lower performance. 
One explanation is that the edge slice usually only contains a small portion of the object of interest, making it hard for SAM 2 to propagate further. 
When annotating two more slices (F-Mode3), we can observe a significant improvement only when providing the ground truth mask, and a slight improvement when providing any types of prompts.

\begin{figure*}[t]
    \centering
    \includegraphics[width=\linewidth]{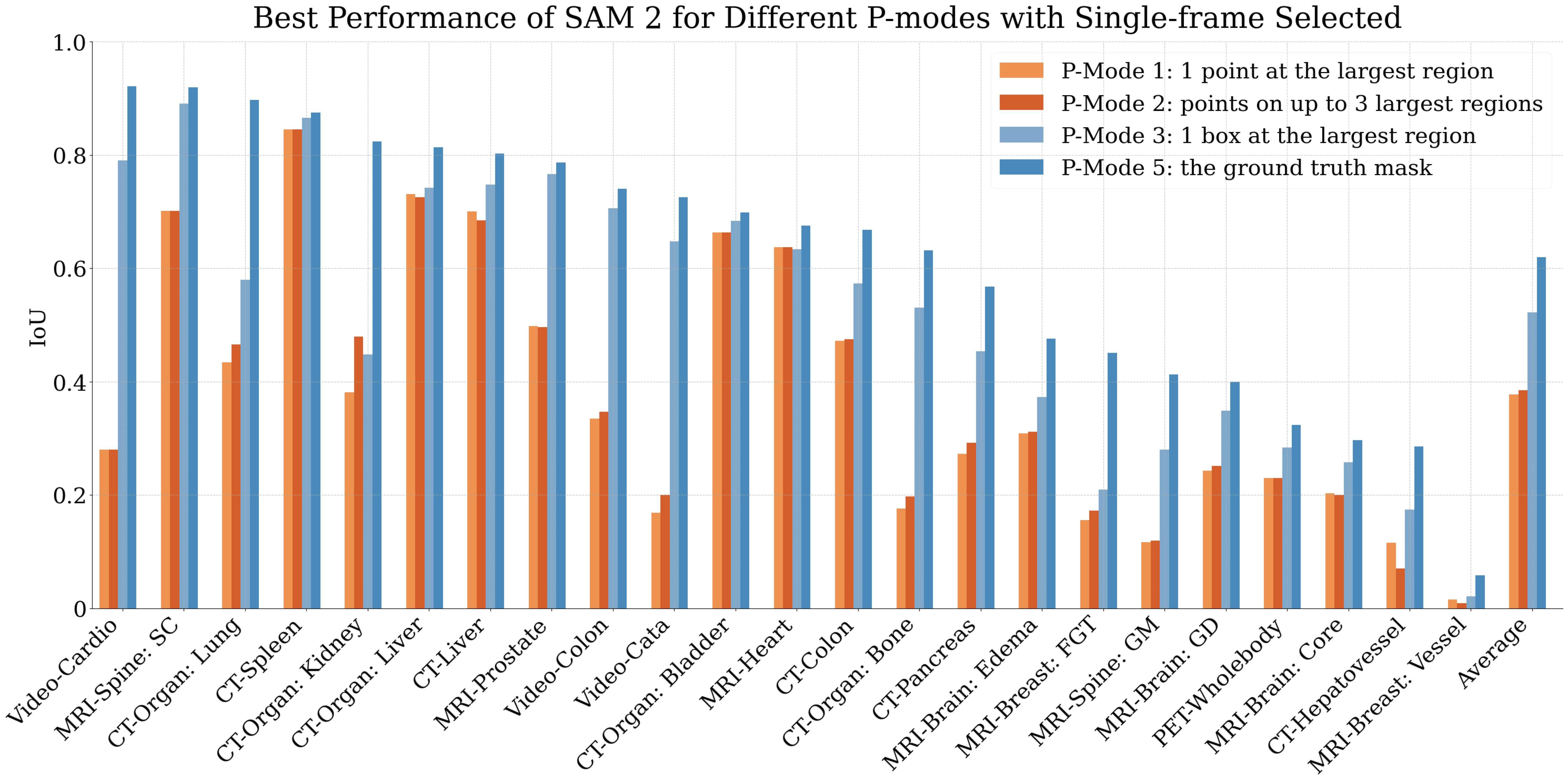}
    \caption{The best performance of SAM 2 under \threedeval when a \textbf{single} frame is selected. 
    Four prompt modes are considered, with results ranked in descending order based on P-Mode 4.}
    \label{fig:sam2_3d_singleframe}
\end{figure*}

\subsubsection{Impact of Prompt Modes}
\begin{figure*}[t]
    \centering
    \includegraphics[width=\linewidth]{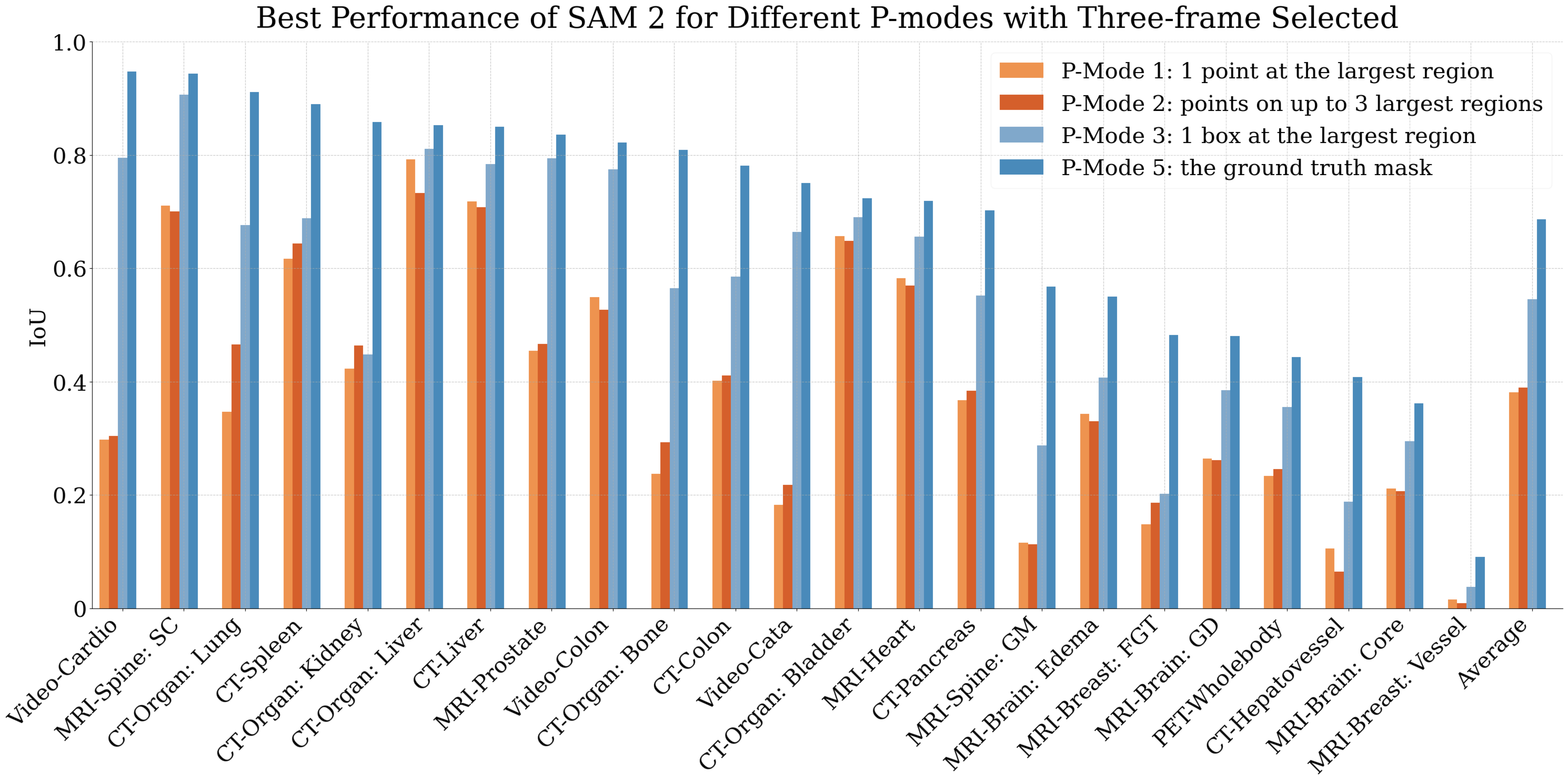}
    \caption{The best performance of SAM 2 under \threedeval when \textbf{three} frames is selected. 
    Four prompt modes are considered, with results ranked in descending order based on P-Mode 4.}
    \label{fig:sam2_3d_multiframe}
\end{figure*}

For different prompt modes (as illustrated by different rows in Figure \ref{fig:result_enumerate}), we find that providing more complicated prompts helps. 
Specifically, the best performances for providing 1 point at the largest region, points on up to 3 largest regions, 1 box at the largest region, and the ground truth mask on a single slice are 0.3778, 0.3851, 0.5222, and 0.6198. Increasing the number of annotated slices to three results in differences of +0.004, +0.004, +0.024, and +0.067 respectively. 
Figure \ref{fig:sam2_3d_singleframe} and Figure \ref{fig:sam2_3d_multiframe} display the performance of the best setting for the single-slice and multiple-slice scenarios on individual datasets respectively. 
This rank of different prompt modes agrees with SAM 2 under the \twodeval, suggesting that the quality of the initial slice prediction determines the segmentation performance of the entire volume. 
Additionally, when comparing the results of putting 1 point on 3 uniform slices vs. 3 points on a single slice, we find that they lead to similar performance. 

The results of using ground truth masks further allow us to investigate prediction changes during propagation; it also mimics one usage where users annotate the current slice and want SAM 2 to predict the rest. Despite there being consistent improvements over using predicted masks, using ground truth masks does not yield a high IoU on average (0.6198 on average). We hypothesize that there are two reasons for this behavior: (1) SAM 2 was trained on a frame width of 8 frames, and thus the memory attention is most effective when the propagated frame is within 8 frames of the annotated frame. Medical volumes range from tens to hundreds of slices. Consequentially, the performance will be significantly affected when there are more slices per volumes. (2) SAM 2's primary objective is to segment any videos. Although this function can be transferred seamlessly between video segmentation and 3D segmentation, videos and medical volumes can differ significantly. Namely, the change between consecutive slices can be large for medical volumes due to the nature of medical imaging technology. In this case, we believe that using SAM 2 for 3D medical imaging segmentation can be further improved.

\begin{table}[t]
    \centering
    \begin{tabular}{|l|c|c|}
       \toprule
       P-Mode & \# of Slices & Performance \\
       \midrule
       \midrule
       \multirow{3}{3cm}{1: 1 point at the \\[.4\baselineskip] largest region}  & 1 & 0.4297 \\
       & 3 & 0.4214 \\
       & Every & 0.4974 \\
       \midrule
        \multirow{3}{3cm}{2: points on up to \\[.4\baselineskip] 3 largest regions}    & 1 & 0.4361 \\
       & 3 & 0.4376 \\
       & Every & 0.5211 \\
       \midrule
        \multirow{3}{3cm}{3: 1 box at the \\[.4\baselineskip]largest region}    & 1 & 0.5354 \\
       & 3 & 0.5600 \\
       & Every & 0.6202 \\
       \midrule
       \multirow{3}{*}{5: the ground truth mask}    & 1 & 0.6310 \\
       & 3 & 0.7050 \\
       & Every & 1.00 \\
       \bottomrule
    \end{tabular}
    \caption{The average performance across all datasets used by both \twodeval and \threedeval. Different prompt modes with annotating 1, 3, and every slice of the volume are considered.}
    \label{tab:slice_based_avg}
\end{table}

\subsection{Comparison between SAM 2 under 3D Segmentation and Others} \label{experiment:sam2_and_others}
In this section, we aim to answer two interesting research questions for SAM 2: (1) What is the difference between providing prompts at the \textbf{volume level} vs. \textbf{image level}? (2) Is SAM 2 more effective than fine-tuning SAM into the 3D setting? 

\begin{figure}[t]
    \centering
    \includegraphics[width=\linewidth]{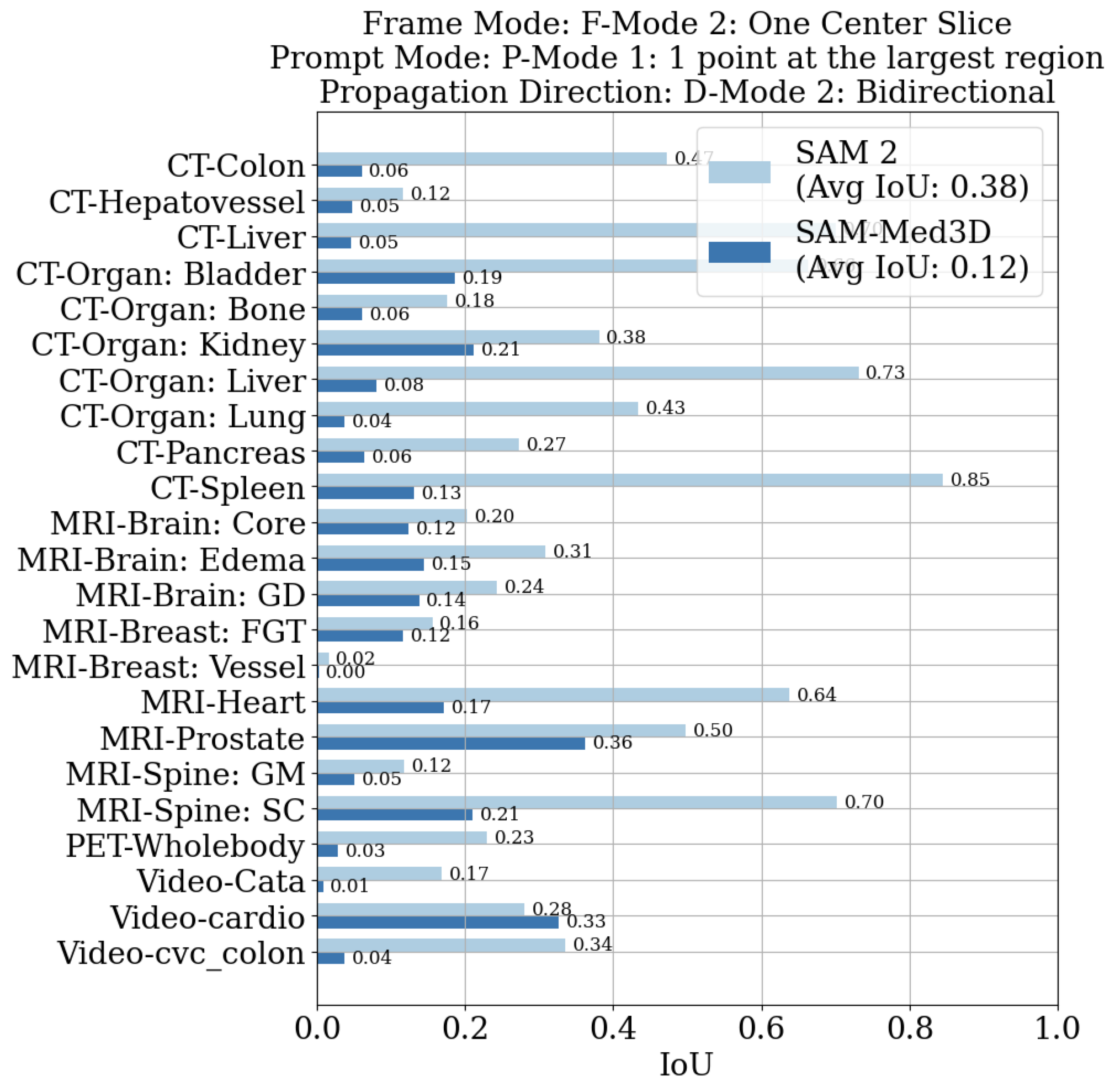}
    \caption{Comparison between SAM 2 and SAM-Med3D on P-Mode 1, F-Mode 2 across 23 3D medical imaging tasks.}
    \label{fig:sammed3d}
\end{figure}

\begin{figure*}
    \centering
    \includegraphics[width=0.9\linewidth]{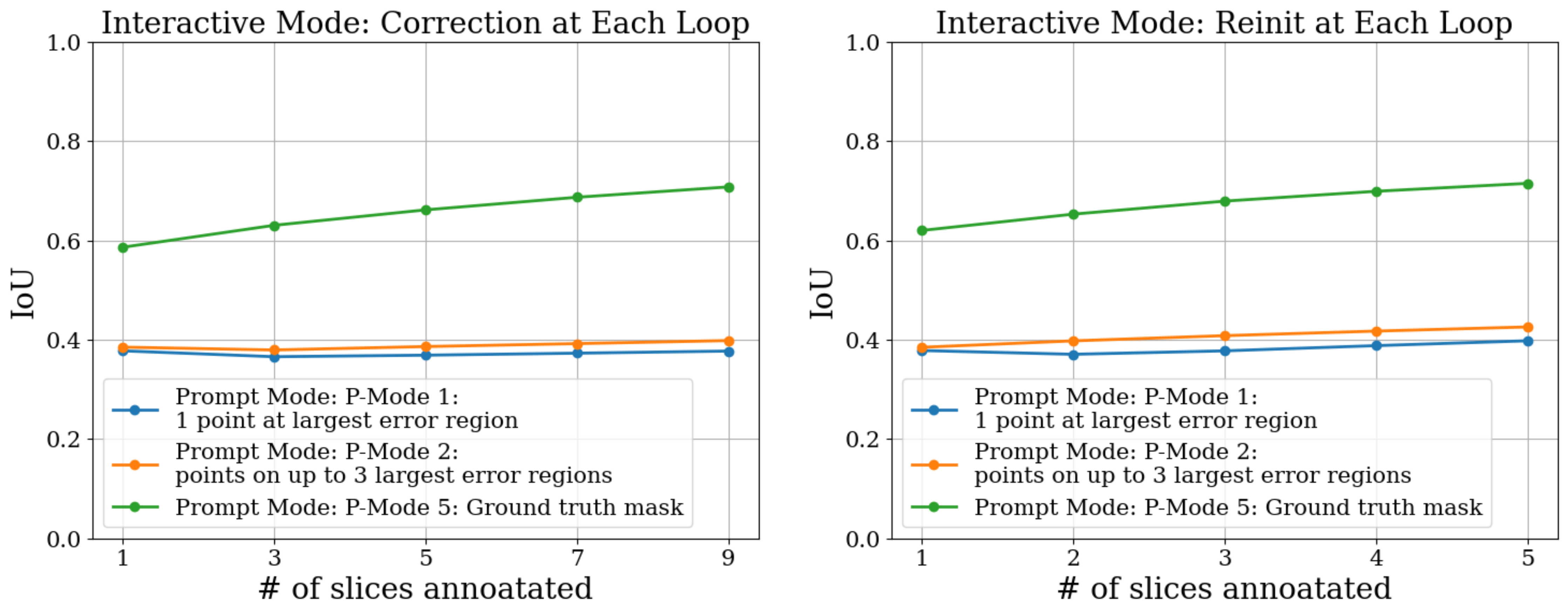}
    \caption{Performance of two types of interactive prompting segmentation.}
    \label{fig:interact_result}
\end{figure*}

\begin{figure*}
    \centering
    \includegraphics[width=0.9\linewidth]{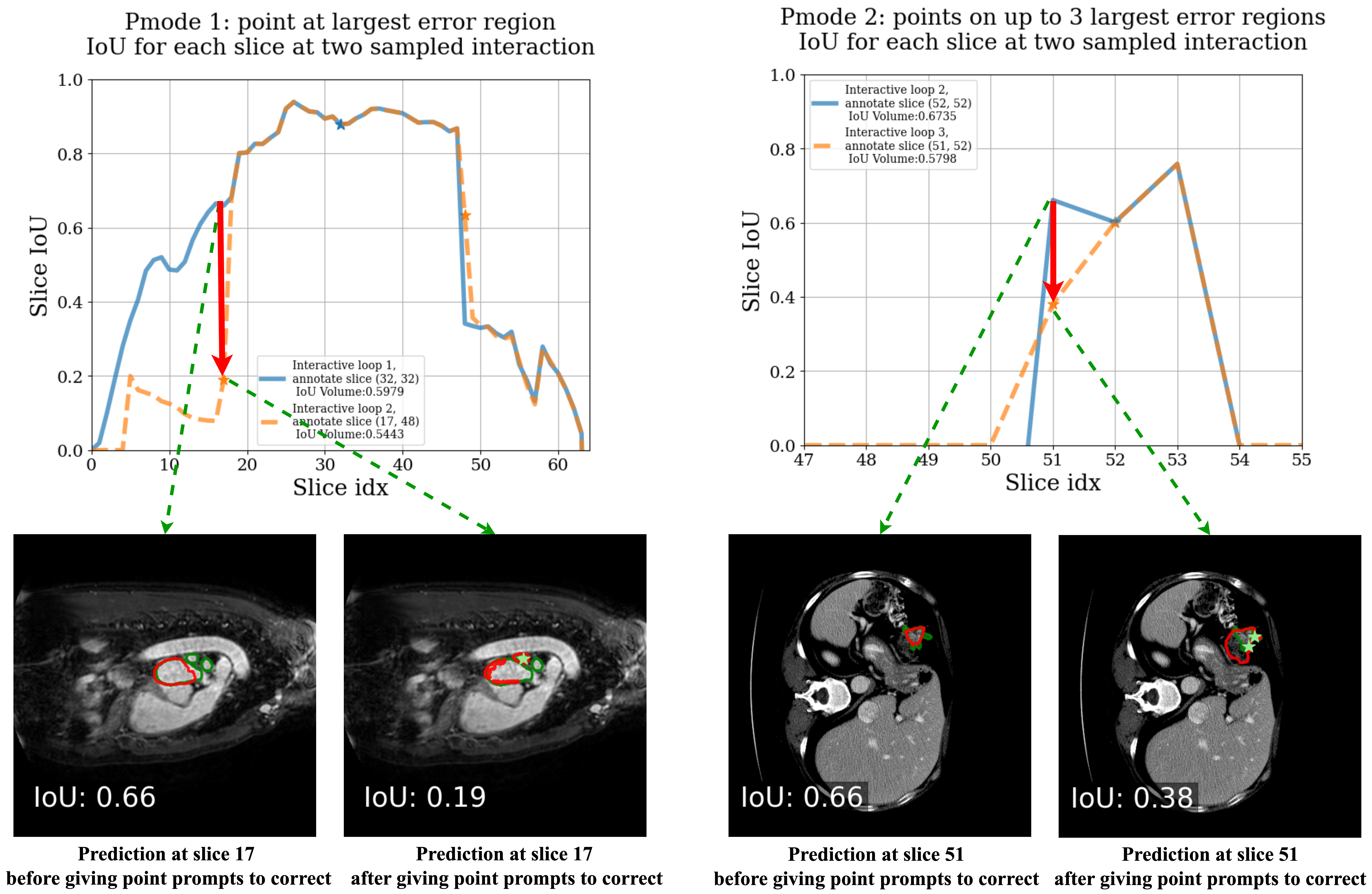}
    \caption{Examples of the score distribution of a single volume at different interaction loops. The predictions of the same slice at different loops are also presented, where the red/green contours indicate prediction/ground truth mask.}
    \label{fig:interact_exam}
\end{figure*}

\subsubsection{The effect of Single-frame and Multi-frame Evaluation}
To address the first question, we can compare the average performance when providing prompts to one slice, three slices, and every slice. The results are shown in Table \ref{tab:slice_based_avg}. To make the results comparable, we compute the average performance across the datasets used by both \twodeval and \threedeval.
Although we have found the optimal setting for each prompt mode in the previous section, there is still a gap between \twodeval and \threedeval. 
Additionally, when comparing different prompt types (point, box, and ground truth mask), we observe that providing box prompts results in the smallest gap between single-frame and multi-frame segmentation performance. This suggests that if a user plans to annotate only a few slices for volume segmentation, using box prompts is the most efficient choice, as it minimizes the performance drop compared to annotating all slices.

Note that the comparison is not fair since prompts are provided on every slice in the single-frame setting, whereas in the multi-frame setting, only up to three prompts are provided to the entire volume, and SAM 2 utilizes the memory bank for predicting slices without prompts.
On the other hand, in the multi-frame setting, the memory bank can utilize knowledge from adjacent slices, which also brings additional knowledge on the third dimension of the volume compared with the single-frame setting.
Based on the gaps observed between single-frame and multi-frame settings, we can conclude that providing prompts on new slices (as human input adds additional information) is still more effective than relying on SAM 2's automatic knowledge transfer between slices.
We hope this discussion can inspire future work, such as more effectively utilizing the memory bank for medical image features, to narrow the gap between the two settings and even exceed the performance of 2D single-frame segmentation.

\subsubsection{Comparison between SAM 2 and SAM fine-tuned to 3D}
To address the second question, we select SAM-Med3D \cite{wang2023sammed3d} as the representative method that fine-tunes SAM \cite{kirillov2023segment} for 3D medical imaging. Since SAM-Med3D only supports point prompts, we compare it to SAM 2's optimal setting when using a single prompt, \textit{i.e.,} P-Mode1, F-Mode2, D-Mode2, and S-Mode2. The results on all datasets are shown in Figure \ref{fig:sammed3d}, in which SAM 2 outperforms SAM-Med3D on nearly all datasets, demonstrating its robustness and effectiveness. One reason for the effectiveness of SAM 2 is that SAM-Med3D receives the full volume as inputs and thus has to reduce the input size significantly to $128 \times 128 \times 128$, while SAM 2 maintains the high-resolution inputs at $1024 \times 1024$ thanks to its propagation strategy. To ensure a fair comparison between the two methods, we resize the input to the required size for SAM-Med3D, instead of using the default center crop.

\subsection{Interactive Segmentation Performance}
In the correction-based algorithm, we identify a slice within a specified range (16 slices) and provide prompts to correct previous predictions. In this setting, we find that adding point prompts at each loop does not improve performance significantly. In P-mode 1, the performance remains unchanged as more slices are annotated, whereas in P-mode 2, the performance increases from 0.3846 to 0.3981 after 9 slices are annotated. Considering there are $8\times 3$ more point prompts provided on each volume, we believe this improvement is minor.
Upon closer examination, as illustrated in Figure \ref{fig:interact_exam}, we discover that providing additional point prompts on slices with prior predictions does not guarantee performance improvement for this slice. 
While the targeted region might be corrected, other regions on the same slice may yield worse predictions, potentially decreasing the slice’s overall prediction accuracy. This unstable correction process can also cause the entire volume’s predictions to become unstable with each propagation loop. 
However, when providing a ground truth (GT) mask at each interactive loop—which can generate better performance on the prompted slice—the average performance across loops gradually improves. 

In the reinitiation-interactive algorithm, we add new prompts to the slice with the lowest IoU in the entire volume and reinitialize the model at each loop. The results are shown in Figure \ref{fig:interact_result} (right). 
Although this strategy does not rely on past memories and thus avoids the instability of correcting previous predictions, we do not observe a significant improvement in performance.
We believe this is because slices with the lowest IoU typically occur at the object's boundary or are particularly challenging, where providing prompts on these slices may not help propagate improvements to other slices and may not even enhance the performance of the targeted slice. Despite the relatively small improvement, we find that this strategy still outperforms the first one. For instance, P-mode 1 shows a slight enhancement, increasing from 0.3782 to 0.3977, rather than remaining unchanged. Additionally, P-mode 2 demonstrates a more noticeable improvement, rising from 0.3846 to 0.4256, even with fewer annotated slices. These observations suggest that in the interactive mode, clearing the memory during interaction is more effective than retaining it for correction. 
When ground truth (P-mode 5) masks are provided for these low-performance slices, the performances at each loop are 0.6198, 0.6526, 0.6789, 0.6987, and 0.7146, respectively.

In both algorithms, regardless of whether memory is cleared or retained, we observe that providing point prompts, using prompt mode 1 (1 point at the largest region) or prompt mode 2 (points on up to 3 largest regions), results in similar performance both between different prompt modes.
This observation aligns with previous non-interactive 3D segmentation findings, as shown in Table \ref{tab:slice_based_avg} for both the 1-point prompt and 3-point prompt modes. Specifically, adding more points to the same slice does not enhance performance, even when selecting new prompts in the most error-prone regions. 
Additionally, based on the observation that providing point prompts during interactive annotation and the previous finding that annotating uniformly spaced slices (F-mode 4) does not improve performance (Table \ref{tab:slice_based_avg}), we conclude that providing point prompts on more frames, whether through uniform selection of slices or by interactively targeting challenging slices, does not yield significant performance gains. 

It is important to note that, although the improvement for the correction-based interactive algorithm appears more significant compared to the reinitialization-based mode, it involves annotating twice as many slices at each loop—one slice during the left propagation and another during the right propagation. Comparing the correction-based mode at the 5th loop and interactive mode at the 3rd loop, which both have 5 slices annotated in total, their performance is quite similar.
Additionally, the different trend of the curve between GT mode and two point modes shows that, when applying interactive annotation strategy, it is very important that the additional slice to be annotated/corrected can provide higher performance than predictions from propagation. An unsatisfactory correction is harmful to the entire volume.


\section{Conclusion}\label{sec:dis}

In this work, we investigate the performance of SAM 2 thoroughly in the medical imaging field. 
In addition to evaluating the 2D segmentation task, we investigate its ability to perform the 3D segmentation task, thanks to SAM 2's ability to segment videos. 
Our findings indicate that under \twodeval, 
\begin{enumerate}
    \item SAM 2 exhibits similar performance to that of SAM.
\end{enumerate} 
Under \threedeval, our observation is that:
\begin{enumerate}
    \item For the multi-mask outputs, selecting the first channel is better than selecting the channel with the largest confidence during propagation when having point prompts.
    \item For the initial frame selection, selecting multiple slices is better than selecting one slice with the cost of providing more prompts, and selecting the center slice tends to be the most cost-effective choice.
    \item For prompt selection, box prompts are more effective than point prompts with a higher cost of human effort. 
    \item Bidirectional propagation, starting from the annotated slice, is a more effective strategy when compared to propagating from beginning to end. 
    \item Interactive segmentation is useful only when the new slice is annotated manually, \textit{i.e.,} providing the ground truth mask, but not through prompts. 
    \item The best average 3D performance of SAM 2 when providing one point prompt, one box prompt, and the ground truth mask to the \textbf{entire volume} is 0.3778, 0.5222, and 0.6198 IoU respectively. 
\end{enumerate}

\clearpage
\bibliographystyle{splncs04}
\bibliography{references}

\newpage
\section{Appendix}
\begin{figure*}
    \centering
    \includegraphics[width=1\linewidth]{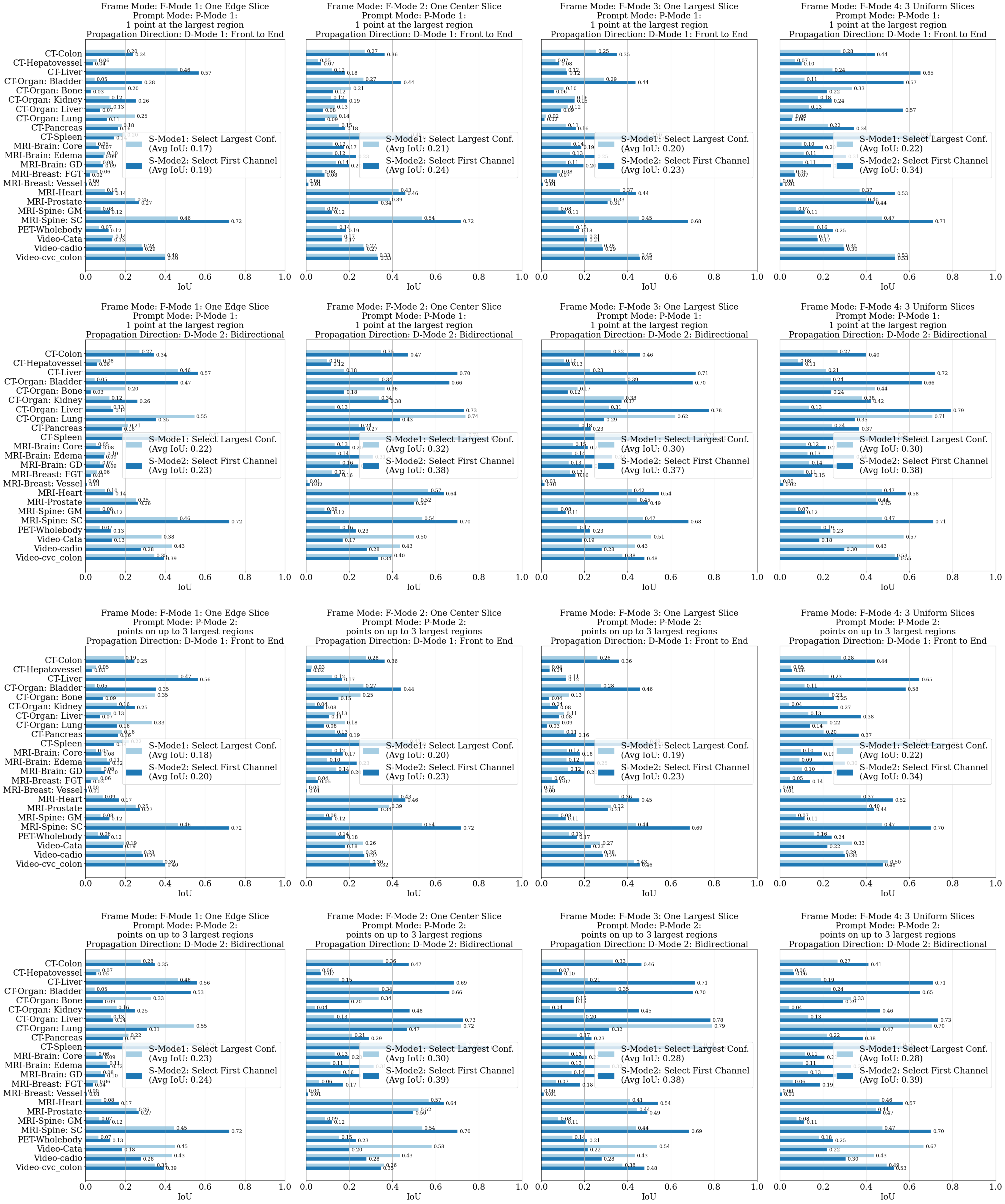}
    \caption{Performance of different choices for different frame choices: F-mode:1,2,3,4; prediction channel choices and two different point prompt modes: P-mode:1 and 2.}
    \label{fig:app_point1}
\end{figure*}

\begin{figure*}
    \centering
    \includegraphics[width=1\linewidth]{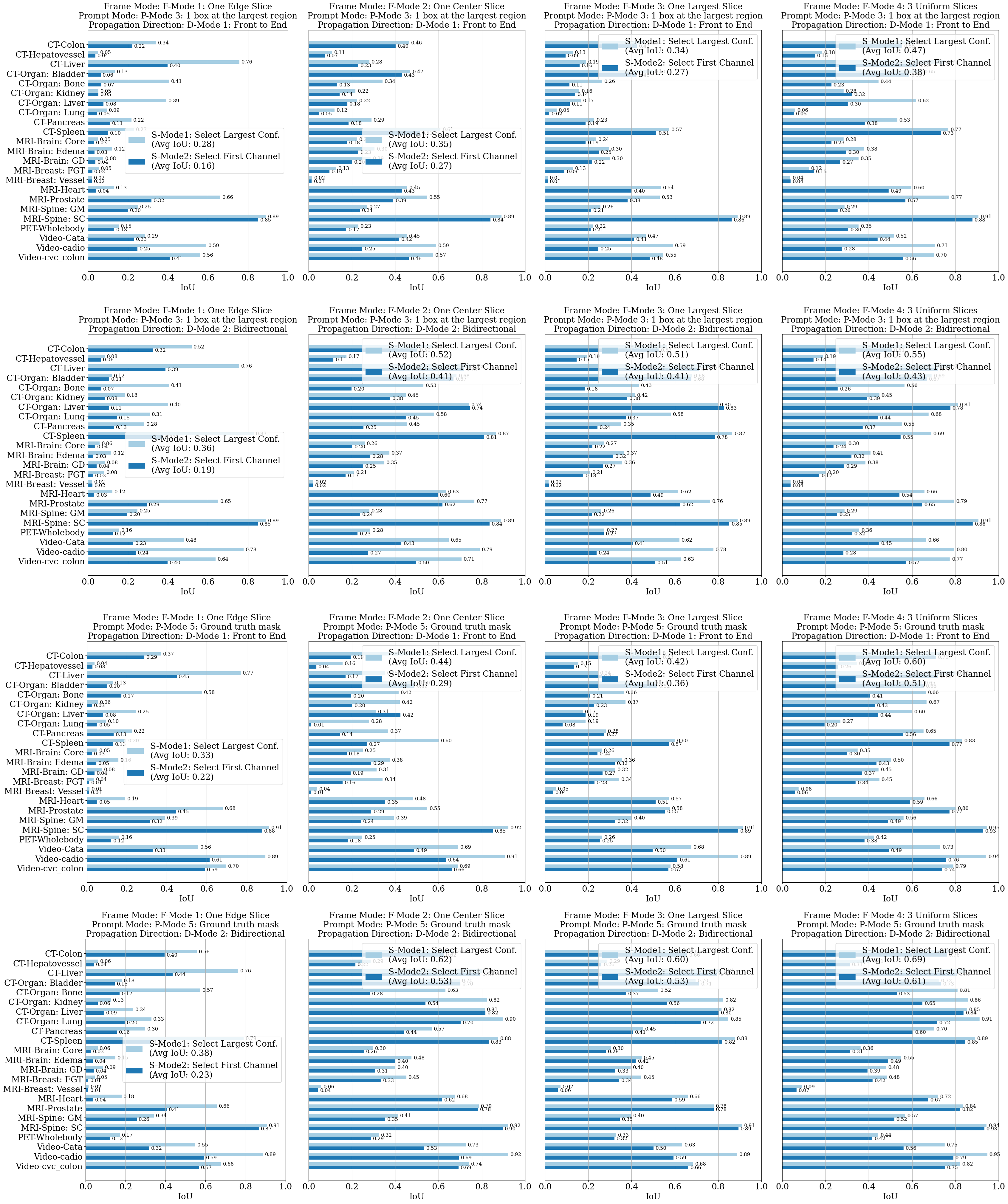}
    \caption{Continual of Figure \ref{fig:app_point1} on P-mode: 3 and 5.}
    \label{fig:app_point2}
\end{figure*}

\begin{figure*}
    \centering
    \includegraphics[width=1\linewidth]{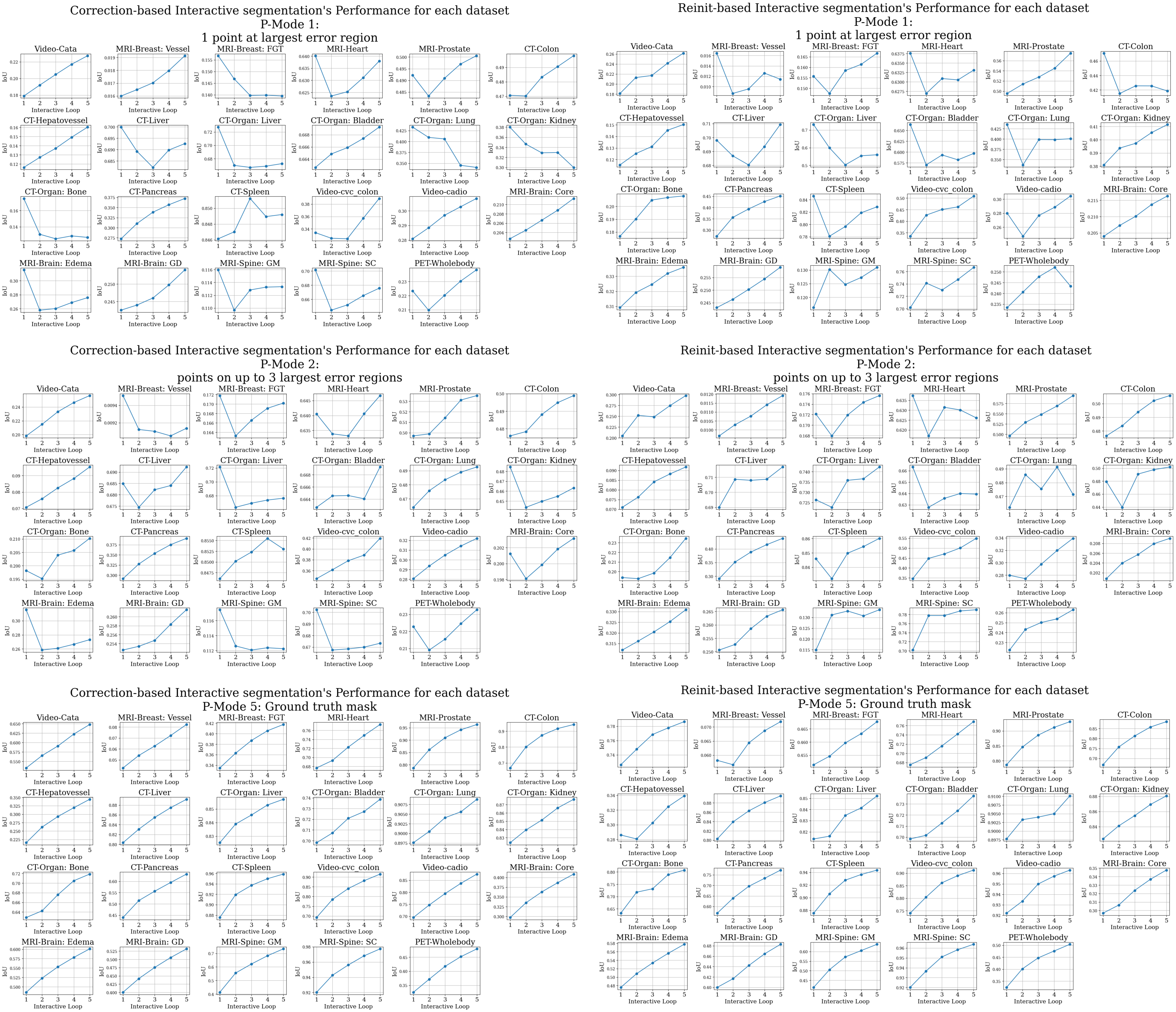}
    \caption{Performance of each dataset under interactive prompting setting, this is the detail performance for Figure \ref{fig:interact_result}.}
    \label{fig:interac_dataset}
\end{figure*}

In the appendix, we present the performance of SAM 2 on individual datasets. Figure \ref{fig:app_point1} and Figure \ref{fig:app_point2} show the performance of different mode combinations introduced in Figure \ref{fig:pipeline}. Figure \ref{fig:interac_dataset} shows the performance of the two interactive algorithms with different prompt modes. 

\end{document}